\documentclass[11pt,a4paper]{article}
\usepackage[hyperref]{emnlp2020}
\usepackage{times}
\usepackage{latexsym}

\usepackage{amsmath}
\usepackage{amsfonts}
\usepackage{bm}
\usepackage{graphicx}
\usepackage{algorithm}
\usepackage{algorithmic}
\usepackage{multirow}

\newcommand*{\algoname}{CR learning}
\newcommand*{\ealm}{\textsf{EALM}}
\newcommand*{\seq}{\textsf{SEQ3}}
\newcommand*{\cmatch}{\textsf{CMatch}}
\newcommand*{\leadn}{\textsf{Lead-N}}
\newcommand*{\orig}{{\boldsymbol{x}}}
\newcommand*{\comp}{{\boldsymbol{y}}}
\newcommand*{\reco}{\boldsymbol{\hat{x}}}
\newcommand*{\lorig}{N}
\newcommand*{\lcomp}{M}

\newcommand*{\zett}{\boldsymbol{z}}
\newcommand*{\act}{{\boldsymbol{a}}}
\newcommand*{\fcomp}{f(\orig, \comp)}
\newcommand*{\freco}{g(\orig, \reco)}
\newcommand*{\arem}{\texttt{Remove}}
\newcommand*{\akep}{\texttt{Keep}}
\newcommand*{\arep}{\texttt{Replace}}

\newcommand*{\crates}{cr_{(t)}}

\newcommand*{\rrates}{rr_{(t)}}
\newcommand*{\creward}{r_{C}}
\newcommand*{\rreward}{r_{R}}

\newcommand*{\breward}{r_{SA}}
\newcommand*{\mask}{\textsc{[mask]}}

\newcommand*{\localenc}{{\boldsymbol{l}}}
\newcommand*{\globalenc}{{\boldsymbol{g}}}
\newcommand*{\freward}{r(s, a,
{\boldsymbol{x}}, {\boldsymbol{y}}, {\boldsymbol{\hat{x}}})}

\newcommand*{\stopwords}{W}

\newcommand{\argmax}{\mathop{\rm arg~max}\limits}
\newcommand*{\dgiga}{\textsc{giga}}
\newcommand*{\dduca}{\textsc{duc3}}
\newcommand*{\dducb}{\textsc{duc4}}

\newcommand*{\lmconv}{LM converter}
\newcommand*{\vsec}{\vspace{-0pt}}
\newcommand*{\vsecu}{\vspace{-0pt}}
\newcommand*{\vsubsec}{\vspace{-0pt}}

\usepackage{amssymb}
\usepackage{pifont}
\newcommand{\cmark}{\textcolor{blue}{\ding{52}}}%
\newcommand{\xmark}{\textcolor{red}{\ding{56}}}%

\usepackage{array}
\newcolumntype{P}[1]{>{\centering\arraybackslash}p{#1}}

\usepackage{tabularx}

\usepackage{url}

\usepackage{microtype}

\aclfinalcopy 
\def\aclpaperid{471} 

\title{Q-learning with Language Model for Edit-based\\ Unsupervised Summarization}

\author{Ryosuke Kohita \\
  IBM Research \\
   \\\And
  Akifumi Wachi \\
  IBM Research \\
  \texttt{\{kohi, akifumi.wachi, yangzhao\}@ibm.com} \\\And
  Yang Zhao \\
  IBM Research \\
  \\\And
  Ryuki Tachibana \\
  IBM Research \\
  \texttt{ryuki@jp.ibm.com}
  
  \\}

\date{}

\begin{document}
\maketitle
\begin{abstract}
Unsupervised methods are promising for abstractive textsummarization in that the parallel corpora is not required. 
However, their performance is still far from being satisfied, therefore research on promising solutions is on-going.  
In this paper, we propose a new approach based on Q-learning with an edit-based summarization. 
The method combines two key modules to form an Editorial Agent and Language Model converter (\ealm). 
The agent predicts edit actions (e.t., delete, keep, and replace), and then the LM converter deterministically generates a summary on the basis of the action signals. 
Q-learning is leveraged to train the agent to produce proper edit actions. 
Experimental results show that \ealm~delivered competitive performance compared with the previous encoder-decoder-based methods, even with truly zero paired data (i.e., no validation set).
Defining the task as Q-learning enables us not only to develop a competitive method but also to make the latest techniques in reinforcement learning available for unsupervised summarization.
We also conduct qualitative analysis, providing insights into future study on unsupervised summarizers.\footnote{Our codes are available at \url{https://github.com/kohilin/ealm}}
\end{abstract}

\section{Introduction}
\label{sec:introduction}
\vsec
Automatic text summarization\footnote{We refer to abstractive summarization in this paper.} is an attractive technique for helping humans to grasp the content of documents effortlessly.
While supervised neural methods have shown good performances~\cite{see-etal-2017-get, zhang-etal-2019-pretraining}, the unsupervised approach is starting to attract interest due to its advantage of not requiring costly parallel corpora.
However, the empirical performance of unsupervised methods is currently behind that of state-of-the-art supervised models~\cite{zhao-etal-2018-language, baziotis-etal-2019-seq}.
Unsupervised text summarization is still developing and is now at the stage where various solutions should be actively explored.

\begin{figure}[t]
    \centering
    \includegraphics[scale=0.90]{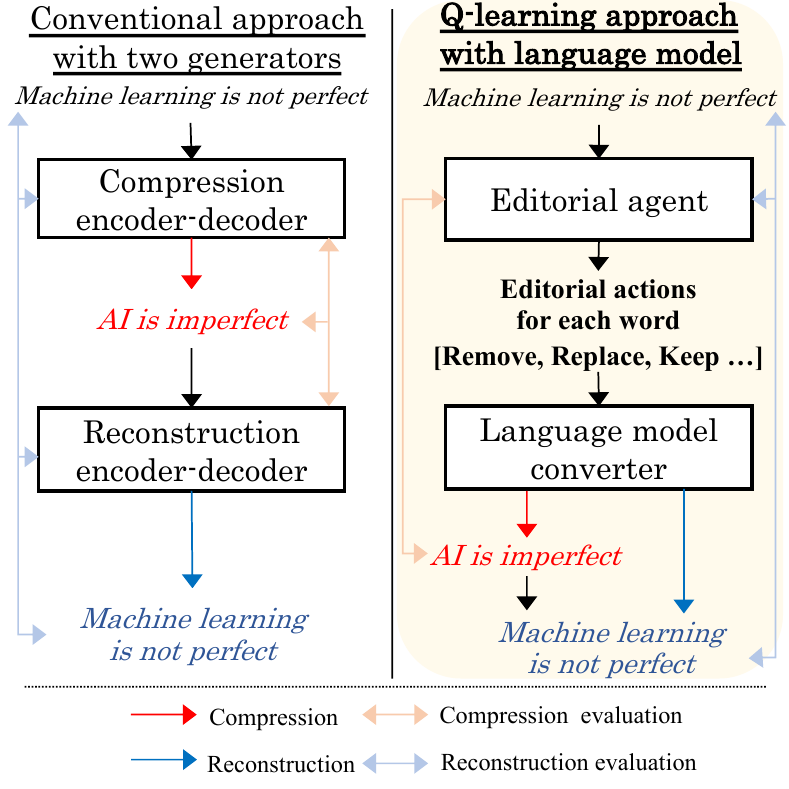}
    \caption{Overview of previous (left) and proposed (right) approaches on \algoname~paradigm.}
    \label{fig:overview}
\end{figure}

One previous unsupervised approach extends neural encoder-decoder modeling to the zero paired data scenario, where a model is trained with a paradigm called compression-reconstruction (CR) learning~\cite{miao-blunsom-2016-language, fevry-phang-2018-unsupervised, zhao-etal-2018-language}.
The mechanism is similar to that of the back-translation~\cite{sennrich-etal-2016-improving}: the model consists of a compressor (i.e., summarizer) and a reconstructor, and they are co-trained so that the reconstructor can recover the original sentence from the summary generated by the compressor~(\citealp{miao-blunsom-2016-language}; the left side of Figure~\ref{fig:overview}).
Experimental results showed that such an unsupervised encoder-decoder-based summarizer is able to learn the mapping from a sentence to a summary without paired data~\cite{baziotis-etal-2019-seq, yang2020ted}.

Reinforcement learning (RL) is also a potential solution for the no paired data situation.
In related fields, for example, there are unsupervised methods for text simplification and text compression with policy-gradient learning~\cite{zhang-lapata-2017-sentence, zhao-etal-2018-language}.
Recent RL techniques take a value-based approach (e.g., Q-learning) such as DQN~\cite{mnih2015human} or the combination of policy and value-based approaches such as Asynchronous Advantage Actor-Critic~\cite{mnih2016asynchronous}.
A critical requirement to leverage a value-based method is a value function that represents the goodness of an action on a given state~\cite{sutton1998introduction}.
We can naturally define the value function by utilizing the CR-learning paradigm, and it makes the latest value-based approaches available for unsupervised text summarization.

In this paper, we propose a new method based on Q-learning and an edit-based summarization~(\citealt{NIPS2019_929tral, malmi-etal-2019-encode}; right side of Figure~\ref{fig:overview}).
The edit-based summarization generates a summary by operating an edit action (e.g., keep, remove, or replace) for each word in the input sentence.
Our method implements the editing process with two modules: 1) an {\bf E}ditorial {\bf A}gent that predicts edit actions, and 2) a {\bf L}anguage {\bf M}model (LM) converter that deterministically decodes a sentence on the basis of action signals, which we call \ealm.
The CR learning is defined on the Q-learning framework to train the agent to predict edit actions that instruct the LM converter to produce a good summary.
Although a vast action space causing sparsity in reward, such as the word generation of an encoder-decoder model, is generally difficult to be learned in RL, our method mitigates this issue thanks to its fewer edit actions and the deterministic decoding of a language model.
Moreover, the formulation by Q-learning enables us to incorporate the latest techniques in RL.

The main contribution of this paper is that we provide a new solution in the form of an unsupervised edit-based summarization leveraging Q-learning and a language model.
Experimental results show that our method achieved a competitive performance with encoder-decoder-based methods even with truly no paired data (i.e., no validation set), and qualitative analysis brings insights as to what current unsupervised models are missing.
Also, the problem formulation on Q-learning enables us to import the latest techniques in RL, which leads to potential improvements in future research.

\vsecu
\section{Task Definition}
\label{sec:problem-statement}
\vsec
We begin by formally defining the problem of unsupervised summarization with the \algoname.
The goal of the task is to produce an informative summary $\comp$ consisting of $\lcomp$ words $y_1, y_2, ..., y_{\lcomp}$ for a given input sentence $\orig$ consisting of $\lorig$ words $x_1, x_2, ..., x_{\lorig}$ where $\lcomp < \lorig$.
The challenge in this task is to learn the transformation from $\orig$ to $\comp$ with only the input sentence $\orig$.

To tackle this, the \algoname~introduces an additional transformation called reconstruction.
The reconstruction requests to reproduce the input sentence $\reco$ from the generated summary $\comp$ where $\reco$ is the reproduced sentence consisting of $\lorig$ words $\hat{x}_1, \hat{x}_2, ..., \hat{x}_{\lorig}$.
In terms of the generated sentences $\comp$ and $\reco$, let $C$ be the compression function and $R$ be the reconstruction function:
%
\[
    \comp = C(\orig, \theta_C) \text{ , } \reco = R(\comp, \theta_R)\text{ ,}
\]
where $\theta_C$ and $\theta_R$ are their respective parameters. Thus, the task can be written as the following optimization problem:
%
\begin{equation}
\label{eq:task-basic-objective}
\theta_C^*, \theta_R^* = \argmax_{\theta_C, \theta_R} \{ \fcomp + \freco \} \nonumber \text{ ,}
\end{equation}
%
where $\fcomp$ and $\freco$ are functions to return a higher value for favorable $\comp$ and $\reco$ in regard to the input sentence $\orig$.
According to the CR learning's hypothesis that the summary should contain enough information to guess the original contents, $\comp$ becomes favorable when the difference between $\orig$ and $\reco$ is smaller while $\comp$ maintains the essential aspects of a summary (e.g., shortness, fluency).

\vsecu
\section{Previous Method}
\label{sec:previous-method}
\vsec
The previous approaches use a generative encoder-decoder model (\citealt{Sutskever:2014:SSL:2969033.2969173}), for the compression and reconstruction functions \cite{miao-blunsom-2016-language, fevry-phang-2018-unsupervised, wang-lee-2018-learning, baziotis-etal-2019-seq}.
%
%
Although the objective functions and implementation details differ depending on the study, the underlying motivation entails the same hypothesis as the \algoname.
For example, \citet{baziotis-etal-2019-seq} introduced four objective functions --- discrepancy of $\comp$ from a pre-trained language model, topical distance of $\orig$ and $\comp$, and the length of $\comp$ and probability difference of $x_i$ and $\hat{x}_i$ --- where the former threes can be regarded as $\fcomp$ and the final one as $\freco$.

While such an encoder-decoder model has performed well on many generation tasks, it suffers from inherent difficulties related to repetition \cite{see-etal-2017-get}, length control \cite{kikuchi-etal-2016-controlling}, and exposure bias \cite{DBLP:journals/corr/RanzatoCAZ15}.
It also runs into convergence problems when co-training multiple generators~\cite{NIPS2016_6125}.

\vsecu
\section{Proposed Method}
\label{sec:proposed-method}
\vsec
\begin{figure*}
    \centering
    \includegraphics[scale=0.47]{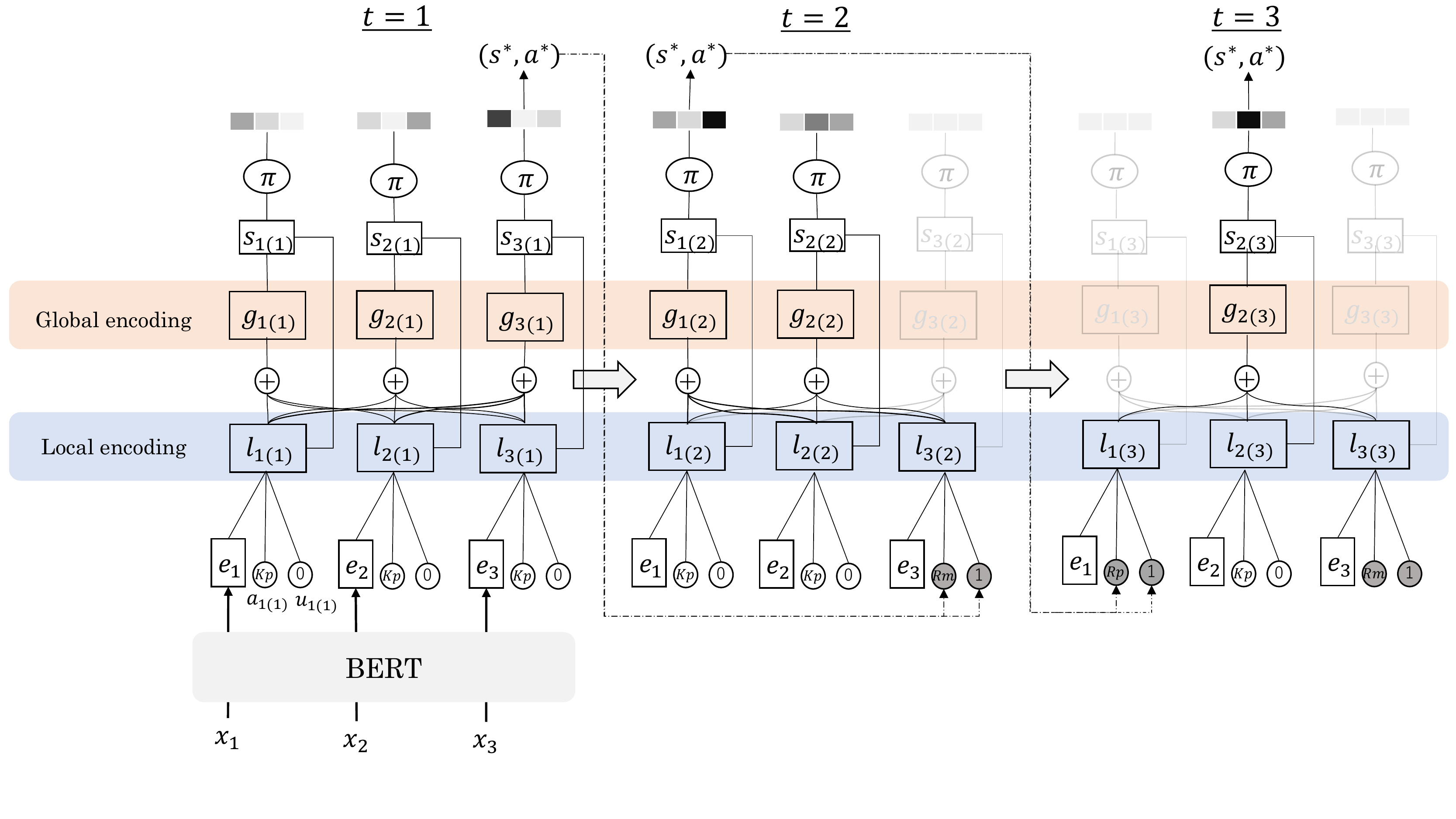}
    \caption{Algorithmic visualization of iterative action prediction}
    \label{fig:algorithm}
\end{figure*}
Our proposed method, which we call \ealm, consists of two essential modules: the editorial agent and the \lmconv.
The agent sends action signals (keep, remove, or replace each word in a sentence) to the conveter, which then deterministically transforms the input sentence according to the signals.
We train the agent to find action signals so that the \lmconv~produces sentences demanded by the \algoname.
In the following sections, we first share the background of Q-learning ($\S$\ref{sec:preliminaries}) and then present how to put the task and our approach on the Q-learning framework ($\S$\ref{sec:uts-as-qlearning}).
We next explain the core algorithmic details ($\S$\ref{sec:algorithm}) and finish with explanations about training and inference ($\S$\ref{sec:training-infernece}).

\subsection{Preliminaries}
\label{sec:preliminaries}
\vsubsec
Q-learning is a popular approach in RL as represented by Deep Q-Networks (DQN, \citealt{mnih2015human}).
Q-learning leverages an action-value function to estimate the {\it value} of a pair of state and action with respect to a policy $\pi$.
The action-value function (i.e., Q-function) is represented as the expected reward for the state-action pair:
%
\[
    Q^\pi(s, a) = \mathbb{E}\left[\sum_{t=0}^{\infty} \gamma^{t} r(s_t, a_t) \mid s_0=s, a_0=a \right],
\]
%
where $s$ is a state, $a$ is an action, $r$ is a reward function for the state-action pair, and $\gamma$ is the discount factor.
Hence, to solve a text summarization task via Q-learning, we first need to appropriately define the state, action, and reward function.

\subsection{Unsupervised Edit-based Summarization with Q-learning}
\label{sec:uts-as-qlearning}
In our approach, given an input sentence $\orig$, we define a state $s_i$ in regard to each word $x_i$.
An action $a_i$ for the state $s_i$ is chosen from among the three options, $\mathcal{A} = \{\arem, \akep, \arep\}$.
The goal of the editorial agent is to provide the optimal action sequence, $\act = \{a_0, a_1, \ldots, a_{\lorig}\}$, by iteratively making action decisions on each word ($\S$\ref{sec:cascaded-action-prediction}).
To obtain $\comp$ and $\reco$, we propose a deterministic transformation algorithm based on $\act$ and the \lmconv~($\S$\ref{sec:deterministic-transformation-via-mlm}). 
Finally, we define the reward function $r$ to evaluate the action and action sequence in terms of the produced sentences $\comp$ and $\reco$ ($\S$\ref{sec:stepwise-reward-computation}).
The reward function is designed to align with the \algoname~paradigm and leads the agent into bringing the action sequence that generates an appropriate $\comp$ and $\reco$.

\subsection{Algorithms}
\label{sec:algorithm}
\vsubsec
In this section, we describe three principle algorithms: 1) how to create $s_i$ and to predict $a_i$, 2) how to generate $\comp$ and $\reco$ by means of $\act$ and the \lmconv, and 3) how to compute the reward.

\subsubsection{Iterative Action Prediction}
\label{sec:cascaded-action-prediction}
The overall flow of iteratively predicting an action for each word is shown in Figure~\ref{fig:algorithm}.
The agent predicts an action for a state (i.e., a word) one by one, so we call one prediction a {\it step} and express it with a subscription $(t)$.
For example, $s_{i(t)}$ and $a_{i(t)}$ respectively denote the state and action for $x_{i}$ at the $t$-th step.
Note that $a_{i(t)}$ has a predicted action if the agent has already done the prediction on $x_i$ by the $t$-th step, otherwise $a_{i(t)}$ is $\akep$.
Also, we prepare a Boolean vector $\bm{u}_{(t)}$ of length $N$ representing the prediction statuses at the $t$-th step; $u_{i(t)}$ is 1 if the prediction on the $i$-th word has been finished, otherwise, 0.
The order to predict an action is determined by Q-values.
Let $s^*$ and $a^*$ be a state-action pair to be operated next, which comes from the maximum Q-values over unoperated states:
%
\[
    s^*, a^* = \argmax_{s\in \mathcal{S}', a \in \mathcal{A}} Q(s, a),
\]
%
where $\mathcal{S}'$ is defined as $\mathcal{S}' = (\forall_i) \{s_{i(t)} \mid u_{i(t)} = 0 \}$.
The agent then reiterates the predictions until it finishes determining an action on all words.
By defining the state in regard to a single word instead of a whole sentence and asking the agent to determine the prediction order, we can handle variable sentence lengths in natural language.
Note that this is not a left-to-right process; the agent conducts the prediction in the order of ``confidence".

Next we explain how to encode $s_{i(t)}$.
To send the agent contextual information, such as the previous decisions, the prediction statuses, and the whole sentence, we dynamically create a state $s_{i(t)}$ with a concatenation of two encodings; {\it local encoding} $\localenc_{i(t)}$ and {\it global encoding} $\globalenc_{i(t)}$
%
\[
    s_{i(t)} = [\localenc_{i(t)}; \globalenc_{i(t)}]\text{.}
\]
To create the two encodings, first, we map $x_i$ to a fixed-sized vector ${\bm e}_i$ with an arbitrary encoder (we use BERT; \citealt{bert-paper}), and $\bm{e}_i$ is repeatedly used throughout the process regardless of the steps.
Then, we define the local encoding as
\[
\localenc_{i(t)} = \bm{e}_i + \bm{b}^{a_{i(t)}} + \bm{b}^{u_{i(t)}}\text{, }
\]
where $\bm{b}^{a_{i(t)}}$ and $\bm{b}^{u_{i(t)}}$ are learnable bias vectors for the action and prediction status of the $i$-th word, respectively.
%
Next, we create the global encoding in a self-attention fashion as
\[
\globalenc_{i(t)} = \sum_{j} w_{j(t)} \localenc_{j(t)}\text{, }
\]
where $w_{j(t)}$ is computed with ReLU:\footnote{We used $\text{ReLU}(\cdot)$ instead of the conventional $\exp(\cdot)$ because $\exp(\cdot)$ caused the exploding gradient in our case.}
\[
w_{j(t)} = \frac
    {
        \text{ReLU}(\localenc_{i(t)} \cdot \localenc_{j(t)})
    }
    {
        \sum_{k}\text{ReLU}(\localenc_{i(t)} \cdot \localenc_{k(t)})\text{.}
    }
\]
Thanks to the bias terms in $\localenc_{i(t)}$ and the self-attention in $\globalenc_{i(t)}$, $s_{i(t)}$ is aware of the previous decisions for each word and the interactions between those decisions.
In addition, BERT encoding $\bm{e}_i$ enables us to take a whole sentence into account.

\subsubsection{Deterministic Decoding by Language Model with Action Signals}
\label{sec:deterministic-transformation-via-mlm}
In this section, we explain how to compress and reconstruct sentences in a deterministic manner with the \lmconv.
For the \lmconv, we use BERT~\cite{bert-paper} which is a masked language model (MLM) trained to predict ``masked" portions in a sentence.
MLM can estimate the probability distribution of $i$-th word $x_i$ in a sentence as $p(x \mid \orig_{\backslash i})$
%
%
where $\orig_{\backslash i}$ is the same as $\orig$ except that it has a mask at the $i$-th position ($\langle ..., x_{i-1}, \mask, x_{i+1}, ... \rangle$; \citealt{wang-cho-2019-bert}).
$L(\orig_{\backslash i})$ denotes a function to return a word with the highest probability for the $i$-th position.

\begin{figure}[t]
    \centering
    \includegraphics[scale=0.85]{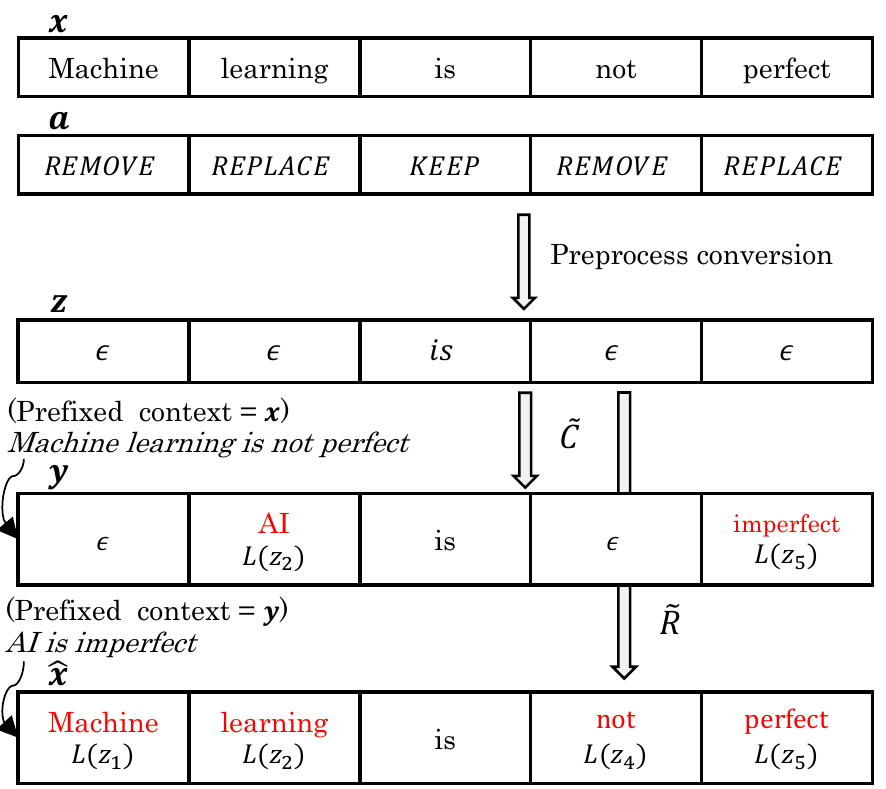}
    \caption{Deterministic compression and reconstruction with masked language model}
    \label{fig:deterministic-comp-recon}
\end{figure}

The procedure to obtain $\comp$ and $\reco$ by using $\act$ and MLM is shown in Figure \ref{fig:deterministic-comp-recon}.
First, we convert $\orig$ to a skeleton sequence $\zett$ consisting of $N$ tokens $z_1, z_2, ..., z_N$ where $z_i$ is $x_i$ if $a_i$ is $\akep$, otherwise a null token $\epsilon$.
%
%
We then define our compression and reconstruction functions $\tilde{C}$ and $\tilde{R}$ as
\begin{align}
\label{eq:proposed-compression-reconstruction}
    y_i &= \tilde{C}(\zett, a_i, L) \nonumber \\
        &= 
        \begin{cases}
            L(\zett_{\backslash i})& (a_i = \arep) \\ \nonumber
            z_i & (\text{otherwise})\text{, } \\ 
        \end{cases}\\ \nonumber
    \hat{x}_i &= \tilde{R}(\zett, a_i, L) \nonumber \\
        &= 
        \begin{cases}
            L(\zett_{\backslash i})& (a_i \in \{\arem, \arep\}) \\ \nonumber
            z_i & (\text{otherwise})\text{. }
        \end{cases}
\end{align}

A word is predicted only for $\epsilon$ given by $\arep$ in compression, but it does so for all $\epsilon$ in reconstruction.
Also, we set the original sentence as a prefixed context, which comes from $\orig$ in compression and $\comp$ in reconstruction, to make MLM aware of a former meaning.
An example is shown in Figure \ref{fig:deterministic-comp-recon}, where MLM receives ``{\it Machine learning is not perfect .} \mask ~{\it is} \mask~." as the compression input and predicts words for the \mask s.    
If there are multiple masks, we conduct the prediction in an autoregressive fashion (see Appendix \ref{sec:autoregressive-prediction-with-MLM}).
Note that while any language model can be used for the \lmconv, MLM is advantageous because it utilizes before and after contexts, and there is no restriction on looking ahead at upcoming words.

\subsubsection{Stepwise Reward Computation}
\label{sec:stepwise-reward-computation}
In this section, we explain the reward computation of the chosen action by referring to $\comp$ and $\reco$.

As stated in  $\S$\ref{sec:cascaded-action-prediction}, we have an action sequence $\act_{(t)}$ for every step $t$.
When we apply $\tilde{C}$ and $\tilde{R}$ to all the $\act_{(t)}$, we can obtain a list of tuples $(s, a, r, \orig, \comp, \reco)_{(t)}$.
A tuple --- let us say, {\it experience} --- enables us to evaluate a state-action pair with respect to a single transition.
In this section, we propose three techniques --- {\bf step reward}, {\bf violation penalty}, and {\bf summary assessment} --- to evaluate the agent's behavior with the stepwise experiences.
Refer to Table \ref{tab:reward-computation} to see how these work in reward computation with an actual example.

Before moving on to the details, let us define two important notions throughout this section, compression rate ($cr$) and reconstruction rate ($rr$):
%
\[
    \crates = 1- \frac{|\comp_{(t)}|}{|\orig|}\text{,  } \nonumber 
    \rrates = \frac{|\{i \mid x_i = \hat{x}_{i(t)} \}|}{|\orig|}. \nonumber
\]

The CR learning assumes that the higher values of $cr$ and $rr$ are better.
We use these for calculating rewards and pruning experiences.

\paragraph{Step Reward.}
The task of the agent is to produce an action sequence with which the \lmconv~generates an appropriately compressed sentence while keeping the reconstruction successful.
As such, we define the reward function $r$ as
%
\[
    \freward = r_{SR} + \breward,
\]
%
where $r_{SR}$ is the step reward that are designed to encourage the agent to improve the compression and reconstruction rate, respectively.
$\breward$ is an additional score from the qualitative assessment of $\comp$, which we explain later. 
Returning to the step reward $r_{SR}$, it is a multiplication of $\creward$ and $\rreward$ defined as
%
\[
    r_{SR} = \creward \times \rreward \text{,}
\]
\[
    \creward = 1 - \frac{|\comp_{(t)}|}{|\comp_{(t-1)}|}\text{, }
    \rreward = 
    \begin{cases}
        1 & (\rrates > \tau_{(t)}) \\
        -1 & (\text{otherwise})
    \end{cases}, \nonumber
\]
%
where $\tau_{(t)}$ is a minimum requirement for the reconstruction rate at the $t$-th step and is defined as $\tau_{(t)} = 1- t \frac{1-\tau}{\lorig}$ with the hyperparameter $\tau \in [0, 1]$.
If we set $\tau = 1$ that requests perfect reconstruction, then $\tau_{(t)} = 1$ regardless of $t$.
However, we need to forgive reconstruction failure to some extent because of the information loss in compression, and $\tau$ adjusts the allowed number of failures.
For example, $\tau = 0.5$ requests the model to recover at least half of the original sentence correctly.

Let us describe the behavior of the step reward $r_{SR}$.
First, the reward is 0 when the agent chooses $\akep$ or $\arep$ because $\creward = 0$ due to there being no change in the length of $\comp$.
Second, the reward gets a positive value when the agent chooses $\arem$ and satisfies the requirement for the reconstruction rate ($\rrates>\tau_{(t)}$).
Third, the reward gets a negative value when the agent chooses $\arem$, but the reconstruction rate is less than the requirement.
In short, the step reward recommends $\arem$ as long as the agent can recover the original word, and otherwise, $\akep$ or $\arep$.

\paragraph{Violation Penalty.}
\label{sec:early-update}
\begin{figure}
    \centering
    \includegraphics{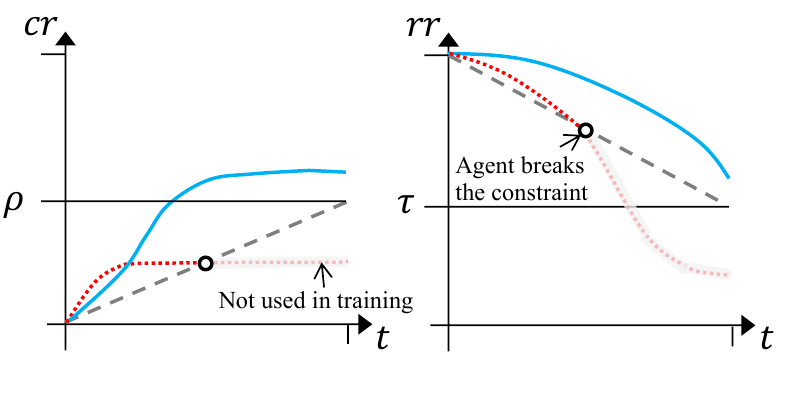}
    \caption{Violation penalty for compression (left) and reconstruction (right). The x-axis is step and the y-axis is each ratio. The horizontal lines in the middle are $\rho$ and $\tau$, and the dashed lines represent $\rho_{(t)}$ and $\tau_{(t)}$. The circles represent a step where the agent breaks the constraints.}
    \label{fig:early-update}
\end{figure}

Sequential modeling, including that performed by our agent, essentially suffers from error propagation caused by incorrect predictions at an earlier stage~\cite{collins-roark-2004-incremental}.
The violation penalty mitigates this issue by giving a negative reward to the latest problematic action and excluding experiences after the mistake.

Here, in addition to $\tau$, we introduce the hyperparameter $\rho$, which represents a minimum requirement for the compression rate.
$\rho_{(t)}$ denotes its threshold at the $t$-step defined as $\rho_{(t)} = t \frac{\rho}{\lorig}$, and the agent must satisfy the condition $\crates > \rho_{(t)}$.
As the penalty, we forcibly assign $-1$ reward for the state-action pair at the $T$-th step when the agent breaks either constraint of $\tau_{(T)}$ or $\rho_{(T)}$.
In addition, we ignore experiences from step $(T+1)$ and onward.
If the agent keeps predicting until the end, we define $T = N$.
Figure \ref{fig:early-update} shows how these constraints work for the experience sequence.

\paragraph{Summary Assessment.}
Although the step reward considers the compression and reconstruction ratios, it ignores the critical aspects of the generated summary such as replacement with a shorter synonym and fluency as a sentence.
Here, we explain the $\breward$ mentioned in the previous paragraph and describe how to reflect such qualitative assessments to the reward given to the agent.

As the essential properties for $\comp$, we take three perspectives into account: informativeness, shortness, and fluency.
The informativeness refers to how much $\comp$ retains the original meaning of $\orig$, and the shortness and fluency are self-explanatory.
To reflect these perspectives onto the agent's decision, we define $\breward$ as
\begin{equation*}
    \begin{split}
        \breward = \frac{T}{\lorig} \cdot 
        &
        [ cr_{(T)} \times rr_{(T)} \\
        &+ \alpha \cdot sim(\orig,\comp_{(T)}) + \beta \cdot llh(\comp_{(T)}) ], \\
    \end{split}
\end{equation*}
where $sim$ computes a similarity score of $\orig$ and $\comp$, and $llh$ computes a log-likelihood of $\comp$.
$\alpha$ and $\beta$ are hyperparameters to adjust the importance of $sim$ and $llh$.
In addition to $r_{SR}$, we give $\breward$ to the experiences from the beginning to $T$-th steps as defined in the step reward paragraph.

\begin{table*}[t]
{\small
\begin{tabular}{p{0.08cm}|c|cccccccccc|c|c}
\hline
               t & Action  & \multicolumn{1}{c|}{Type}      & 1   & 2    & 3      & 4  & 5    & \multicolumn{1}{c|}{6}   & $cr$/$\rho_{(t)}$ & $rr$/$\tau_{(t)}$ & $\frac{T}{N}$/$crrr$/$sim$/$llh$  &  $ r_{SR} $ / $\breward$                 & $r$         \\ \hline
\multirow{2}{*}{1} & $\arem$  & \multicolumn{1}{c|}{$\comp_{(1)}$} &  &{\scriptsize the}  &   {\scriptsize force}     & {\scriptsize be} & {\scriptsize with} & \multicolumn{1}{c|}{{\scriptsize you}} & \multirow{2}{*}{.17/.05} & \multirow{2}{*}{1.0/.91}                    & \multirow{2}{*}{-} & \multirow{2}{*}{1.0 / .20}   & \multirow{2}{*}{1.2}  \\
                  & (May) & \multicolumn{1}{c|}{$\reco_{(1)}$} & {\scriptsize May} & {\scriptsize the}  & {\scriptsize force}  & {\scriptsize be} & {\scriptsize with} & \multicolumn{1}{c|}{{\scriptsize you}} &                              &                                                 &                        &                                           &                        \\ \hline
\multirow{2}{*}{2} & $\arem$  & \multicolumn{1}{c|}{$\comp_{(2)}$} &  &      &  {\scriptsize force}      & {\scriptsize be} & {\scriptsize with} & \multicolumn{1}{c|}{{\scriptsize you}} & \multirow{2}{*}{.33/.10} & \multirow{2}{*}{1.0/.83}                    & \multirow{2}{*}{-} & \multirow{2}{*}{1.0 / .20}  & \multirow{2}{*}{1.2}  \\
                  & (the)   & \multicolumn{1}{c|}{$\reco_{(2)}$} & {\scriptsize May} & {\scriptsize the}  & {\scriptsize force}  & {\scriptsize be} & {\scriptsize with} & \multicolumn{1}{c|}{{\scriptsize you}} &                              &                                                 &       &                                           &                        \\ \hline
\multirow{2}{*}{3} & $\arem$  & \multicolumn{1}{c|}{$\comp_{(3)}$} &     &      &        & {\scriptsize be} & {\scriptsize with} & \multicolumn{1}{c|}{{\scriptsize you}} & \multirow{2}{*}{.50/.15} & \multirow{2}{*}{{\color{red}.50/.75}}                    & \multirow{2}{*}{.50/.25/.50 /1.0} & \multirow{2}{*}{{\bf -1.0} / .20}  & \multirow{2}{*}{-.80} \\
                  & (force)   & \multicolumn{1}{c|}{$\reco_{(3)}$} & {\scriptsize I}   & {\scriptsize will} & {\scriptsize always}  & {\scriptsize be} & {\scriptsize with} & \multicolumn{1}{c|}{{\scriptsize you}} &                              &                                                 &                &                                           &                        \\ \hline
4                  & \multicolumn{13}{c}{ }\\ 
5                  & \multicolumn{13}{c}{{\small No experiences due to the violation occurred at the step 3.}}\\ 
6                  & \multicolumn{13}{c}{ }\\ \hline
\end{tabular}
}
\caption{An example of stepwise reward computation. It breaks the reconstruction constraint at the step 3 when removing {\it force}, so $r_{SR} = -1$. $\breward$ is computed at the step 3 by $ 0.5 \times (0.25 + 0.5 \times 0.1 + 1.0 \times 0.1) = 0.20$, and it is used for the step 1 and 2 as well. The settings of hyperparameters are $\tau=0.5$, $\rho=0.3$, $\alpha=0.1$, and $\beta=0.1$.}
\label{tab:reward-computation}
\end{table*}

Let us explain the terms inside the square brackets first. 
The first term, which is the multiplication of $cr_{(T)}$ and $rr_{(T)}$, aims for shortness and informativeness.
It gets a higher value when the agent achieves the right balance of compression and reconstruction.
The second term $sim$ aims to evaluate informativeness brought about by $\arep$.
Concretely, $sim$ returns a semantic similarity score in the range of $[0, 1]$ through the sentence vectors of $\orig$ and $\comp_{(T)}$ rather than just checking exact matches of words.
The last term $llh$ represents fluency via the log-likelihood of $\comp_{(T)}$ given by a pre-trained language model \cite{zhao-etal-2018-language}.
We use BERT for the computation of $sim$ and $llh$ (\citealt{bert-paper, wang-cho-2019-bert}; see Appendix \ref{sec:sim-llh}).
Finally, $T/N$ is the ratio of the number of operated words.
It becomes closer to 1 when the agent is reaching a termination, i.e., finishing the prediction on all words by avoiding the violation penalty, which makes $\breward$ larger.
In contrast, the agent who fails at an earlier stage gets a small value of $\breward$.

\subsection{Training and Inference}
\label{sec:training-infernece}
\vsubsec
\paragraph{Training.}
Leveraging the experiences $(s, a, r, \orig, \comp, \reco)$ in the replay buffer \cite{DBLP:journals/ml/Lin92}, the agent learns the policy for summarizing a sentence $\orig$ within the Q-learning framework. Specifically, we utilize DQN \cite{mnih2015human} to learn the Q-function $Q^*$ corresponding to the optimal policy by minimizing the loss, 
\[
\mathcal{L}(\theta) = \mathbb{E}_{s, a, r, s'} [(Q^*(s, a)-\psi)^2],
\]
where $\psi = r + \gamma \max_{a'} \bar{Q}^*(s', a')$ and $\bar{Q}$ is a target Q-function whose parameters are periodically updated in accordance with the latest network parameters.
During the collection of experiences, RL requires the agent to explore an action on a given state for finding a better policy.
As a unique point in this work, the agent must explore not only the action but also the order to predict. 
For both explorations, we use the $\epsilon$-greedy algorithm \cite{Watkins:1989} that stochastically forces the agent to ignore Q-values and to behave randomly (see Appendix \ref{sec:exploration-of-prediction-order}).

\paragraph{Inference.}
Our modeling that provides $\comp$ and $\reco$ for each step has another advantage in terms of the inference.
For the final output, we use $\comp$ at the $t^*$-th step that achieves the best balance of the compression and reconstruction ratios, where $t^* = {\rm arg~max}_{t} \{\crates + \rrates\}$.
This is based on the trade-off relationship of compression and reconstruction as seen in the precision-recall curve.

\vsecu
\section{Experiment}
\label{sec:experiment}
\vsec

\paragraph{Baselines.}
We compare our proposed approach with three baselines: \leadn, which simply takes the beginning N words as the summary, \seq, a recent encoder-decoder model~\cite{baziotis-etal-2019-seq}, and \cmatch, a new approach without explicit reconstruction learning~\cite{zhou-rush-2019-simple}.
To conduct qualitative analysis on generated summaries, we ran the baselines ourselves with a replicated model for~\seq\footnote{\url{https://github.com/cbaziotis/seq3}. We ran the training script with the same configuration as the original paper except for decreasing the batch size from 128 to 32 due to our GPU limitation. We trained three models and obtained slightly lower scores than the ones reported in the original paper. We report the averaged score among the three models.} and the provided model for \cmatch.\footnote{\url{https://github.com/jzhou316/Unsupervised-Sentence-Summarization}}
Also, we test two types of \seq~models: one tuned with a validation set ($\seq^{+}$) and the other with parameters at the last iteration in the training ($\seq^{-}$).
This is because \ealm~and \cmatch~do not need paired data even for validation.

\paragraph{Proposed method.}
We implemented \ealm~as follows.
The Q-network of the agent consists of a two-layered MLP with 200 units per layer and ReLU.
We used the Adam optimizer with the learning rate of 0.001 and apply gradient clipping by 1.
For the epsilon-greedy strategy, we first set the exploring probability to 0.9 and decay it by multiplying by 0.995 every 100 updates until it reaches the minimum exploration rate of 0.03.
We set the discount factor $\gamma$ to 0.995.
The size of the replay buffer is 2000, and we sample 128 experiences as a batch for one update.
As the final model, we use parameters at a time when the averaged score of reward in the replay buffer is maximum, i.e., our model does not need a validation set.
The hyperparameters of step reward ($\tau$, $\rho$; $\S$\ref{sec:early-update}) are set to 0.5 and 0.3, respectively.
The hyperparameters of summary assessment ($\alpha$, $\beta$; $\S$\ref{sec:early-update}) are both set to 0.1.
We train three models with the same configuration and report their averaged score, as Q-learning inherently contains randomness in training.

\paragraph{Dataset.}
The same as \citet{baziotis-etal-2019-seq}, we train our model on the Gigaword corpus (\dgiga, \citealp{rush-etal-2015-neural}).
However, we used only 30$K$ sentences randomly picked from sentences with less than $50$ words for the training of~\ealm.
This is because the whole data, 3.8$M$ sentences, is too large to expose the agent to different experiences from the same sentence.\footnote{\ealm~can be trained with the large dataset, but it takes long time due to the exploitation and exploration learning strategy of Q-learning. 30K was better in the balance of the required time and the model performance.}
Note that we used the entirety of sentences for the training of the \seq~models.

We followed \citet{baziotis-etal-2019-seq} in the evaluation as well, using the test set consisting of the $\dgiga$ (1897 sentences) and DUC datasets (\dduca~with 624 sentences, \dducb~with 500 sentences; \citealt{Over:2007:DC:1284916.1285157}).

All models follow the same tokenization policy: the default tokenization in $\dgiga$, $\dduca$, and $\dducb$.
Although BERT (which \ealm~uses) has its own vocabulary based on subwords, we do not apply subwording to go along with a single tokenization policy.
Therefore, words not in the BERT vocabulary are interpreted as unknown words, and the ratio of unknown words was around 10\% in $\dgiga$.

\paragraph{Evaluation.}
In our quantitative analysis, we examine the ROUGE scores.\footnote{We used files2rouge (\url{https://github.com/pltrdy/files2rouge}) following \citealp{baziotis-etal-2019-seq}.}
To mitigate the bias to longer sentences in ROUGE calculation, we capped all summaries at the first 75 bytes.
Note that the averaged sentence length of gold summaries after the capping were 8.58, 9.59, and 10.25 for \dgiga~, \dduca, and \dducb, respectively.
Also, we examine sentence length (LEN) and count of new words (NW; number of words that are used in a generated summary but do not appear in the input sentence).
Additionally, we show qualitative comparisons with a manual check of generated summaries.
Although a questionnaire survey is often conducted to assess the deeper quality of summaries such as informativeness and readability, this still hides the exact points of model's strengths and weaknesses.
We consider that specific indications provide insights on future work for the current unsupervised summarizers.
We manually checked more than 200 summaries for each model and each dataset and include a few samples in Appendix (\ref{sec:samples}).

\paragraph{Results.}
\begin{table}[t]
\centering
{\small
\begin{tabular}{c|l|ccc|c|c}
\hline
\multicolumn{2}{c|}{Data \& Model}    & R-1   & R-2   & R-L   & LEN & NW \\ \hline
\multirow{6}{*}{\dgiga}       & L8     & 21.78 & 7.62  & 20.40 & 8.00   & 0     \\
                            & L15    & 24.22 & 8.20  & 22.00 & 15.00  & 0     \\
                            & $\text{S3}^{+}$   & 23.15 & 7.56  & 21.11 & 14.77  & 0.59  \\
                            & $\text{S3}^{-}$ & 22.09 & 6.59  & 20.02 & 14.63  & 1.09  \\
                            & CM & {\bf 26.71} & {\bf 10.12} & {\bf 24.67} & 9.48   & 0.44  \\
                            & EL   & 25.00 & 7.61  & 22.48 & 17.39  & 0.07  \\ \hline
\multirow{6}{*}{\dduca}    & L8     & 18.34 & 5.76  & 16.92 & 8.00   & 0     \\
                            & L15    & 20.94 & {\bf 6.20}  & 18.54 & 15.00  & 0     \\
                            & $\text{S3}^{+}$   & 20.09 & 5.53  & 17.76 & 16.51  & 0.71  \\
                            & $\text{S3}^{-}$ & 19.57 & 5.17  & 17.25 & 16.42  & 1.17  \\
                            & CM & 17.50  & 4.84  & 16.35 & 5.18   & 0.39  \\
                            & EL   & {\bf 21.69} & 5.25  & {\bf 18.88} & 19.61  & 0.02  \\ \hline
\multirow{6}{*}{\dducb}    & L8     & 18.85 & 4.88  & 17.05 & 8.00   & 0     \\
                            & L15    & 22.14 & {\bf 6.25}  & 19.30 & 15.00  & 0     \\
                            & $\text{S3}^{+}$ & 21.69 & 5.87  & 18.81 & 16.81  & 0.59  \\
                            & $\text{S3}^{-}$ & 21.25 & 5.64  & 18.32 & 16.69  & 1.08  \\
                            & CM & 18.62 & 5.60  & 17.16 & 5.26   & 0.36  \\
                            & EL   & {\bf 22.50} & 5.80  & {\bf 19.47} & 20.46  & 0.01  \\ \hline
\end{tabular}
\caption{ROUGE scores, averaged lengths (LEN), and averaged occurrences of new words (NW). L8 and L15 are \leadn. $\text{S3}^{[+-]}$ represent \seq~models. CM is \cmatch~and EL is \ealm. ROUGE scores are computed with summaries capped at the first 75 bytes.}
\label{tab:result}
}
\end{table}

Table \ref{tab:result} lists the results of ROUGE scores, averaged lengths, and averaged counts of new words. 
\ealm~showed a better performance in \dduca~and \dducb~with respect to R-1 and R-L.
In \dgiga, it performed competitively with the baselines.
However, the original length of the generated summaries tended to be longer, and the occurrence of new words was the lowest.

\cmatch~achieved the highest scores of ROUGE and meaningful length in \dgiga.
The scores of R-2 and R-L were superior to others by about two points, which means \cmatch~captured not only salient words but also word co-occurrences.
However, for generating summaries, \cmatch~uses a language model trained with gold summaries in $\dgiga$.
In other words, it may just internally store the probable word distributions in summary sentences on $\dgiga$.
Actually, the results on $\dduca$ and $\dducb$ were not better than those on $\dgiga$.
Even though \cmatch~does not require paired data, it is not practical to collect enough summaries to train a language model for each domain.

\seq~showed a competitive performance with other models, but its scores dropped when no validation set was available.
The requirement of a validation set is a keen disadvantage because creating input-summary pairs comes at a significant human labor cost.

While almost all of the best scores were given by the statistical models, $\mathsf{Lead}$-15 also performed competitively.
This result indicates that unsupervised summarization methods can not yet overcome the trivial baseline.
One significant barrier preventing the progress of unsupervised methods is presumably the difficulty of rephrasing. 
For writing a good summary, condensing a longer expression into a shorter form is essential.
As seen in the NW column in Table~\ref{tab:result}, the number of new words was less than one in $\seq^{+}$, $\cmatch$, and $\ealm$.
The current models tend to operate just by copy-and-paste, which is consistent with the report by \citet{baziotis-etal-2019-seq}.

\begin{table}[t]
{\small 
\begin{tabular}{llp{5.8cm}}
\hline
\multirow{4}{*}{S3$^{+}$}     & \cmark & Grammatical                         \\
                             & \cmark  & Informative \\
                             & \xmark  & Copy words from the top as it is                 \\
                             & \xmark  & Meaningless rephrasing   \\ \hline
\multirow{4}{*}{CM}          & \cmark  & Grammatical \\
                             & \cmark  & Fluent in successful cases (in $\dgiga$) \\
                             & \xmark  & Lack of information (in $\dduca$ and $\dducb$)       \\ 
                             & \xmark  & Too much short (in $\dduca$ and $\dducb$)      \\ \hline
\multirow{4}{*}{EL}          & \cmark  & Select words from the whole input  \\
                             & \cmark  & Contain keywords \\
                             & \xmark  & Less grammatical \\
                             & \xmark  & Lack of rephrasing \\ \hline
\end{tabular}
\caption{Pros (\cmark)  and cons (\xmark) found in the generated summaries of \seq, \cmatch, and \ealm.}
\label{tab:qualitative-analysis}
}
\end{table}

\begin{table*}[]
{\small 
\begin{tabular}{p{2.8cm}|p{2.8cm}|p{2.8cm}|p{2.8cm}|p{2.8cm}}
\hline
\multicolumn{1}{c}{INPUT}                                                                                                                                                                                        & \multicolumn{1}{|c}{Human}                                          & \multicolumn{1}{|c}{\seq}                                                                                             & \multicolumn{1}{|c}{\cmatch}                                           & \multicolumn{1}{|c}{\ealm}                                                                                                                               \\ \hline
japan 's nec corp. and \textbf{UNK} computer corp. of the united states said wednesday they had agreed to join forces in supercomputer sales .                                                                            & nec UNK in computer sales tie-up                                   & japan 's nec corp. and \textcolor{red}{\textit{her}} computer corp. of the united states said                                                  & nec \textcolor{blue}{\textit{agrees}} to join forces in supercomputer sales & nec computer united states said agreed \textcolor{red}{(\textit{to})} join forces in sales                                                                                            \\ \hline
mechanical problems \textbf{that} threaten to shut down the \textbf{astronomical} observations of the hubble space telescope may \textbf{prompt a repair mission} six months earlier than planned to the \$ 1.7 billion spacecraft , nasa officials told congress on wednesday .                                  & Problems may stop Hubble astronomical observations; NASA may \textbf{accelerate repair mission} & mechanical problems that threaten to shut down the \textcolor{red}{\textit{her}} observations of the hubble space telescope threaten                      & \textcolor{red}{\textit{nasa observes}}                                       & \textcolor{red}{\textit{that}} threaten to shut down astronomical observations of space telescope may \textcolor{blue}{\textit{prompt repair mission}} six earlier than planned billion spacecraft \textcolor{red}{\textit{nasa officials told congress on}}           \\ \hline
endeavour 's astronauts connected the first two \textbf{building blocks of the international space station} on sunday , \textbf{creating a seven-story tower in the shuttle cargo bay} .                                                                                                                                                                                       & First 2 \textbf{building blocks of international space station} successfully joined.  & endeavour 's astronauts connected the first two \textcolor{blue}{\textit{building blocks of the international space station}}                                 & endeavour 's astronauts create a shuttle                      & connected first \textcolor{blue}{\textit{building blocks of international space}} on \textcolor{blue}{\textit{creating tower in shuttle bay}}                                                                                      \\\hline
\end{tabular}
}
\caption{Summaries by Human (gold reference), \seq, \cmatch~and \ealm~from $\dgiga$ (top), $\dduca$ (center), and $\dducb$ (buttom).}
\label{tab:main-examples}
\end{table*}

Finally, we manually assess the summaries produced by each model and sum up their pros and cons in Table \ref{tab:qualitative-analysis}.
Also, actual examples are shown in Table~\ref{tab:main-examples}.
First, we found that a summary of $\seq$ was likely to be an exact copy of the input sentence from the top, but it kept sentences grammatical and informative.
Rephrasing by $\seq$ did not meet our expectation in most cases, such as changing a week of the day (e.g., {\it Wednesday} to {\it Thursday}) or a common adjective to a pronoun adjective (e.g., {\it astronomical} to {\it her}).
$\cmatch$ stably generated fluent summaries in $\dgiga$, as seen in the ROUGE scores.
It also generated grammatically correct sentences such as number agreement (e.g., {\it nec agree\underline{s} ...}).
In the DUC datasets, however, meaningless summaries increased, such as containing no important information (e.g., {\it nasa observes}).
Relatedly, \cmatch's summaries on $\dduca$ and $\dducb$ were too short, and we found that more than half of the summaries consisted of less than or qeual to 5 words.
Finally, \ealm's outputs tended to be longer due to containing non-informative portions (e.g., {\it nasa officials told ...}). It was also likely to be ungrammatical due to leaving only a functional word (e.g., {\it mechanical problems \underline{that} threaten ...}) or deleting required prepositions (e.g., {\it ... agreed (to) join ...}).
Those failures resulted in lower readability.
However, \ealm~tried to keep keywords from the whole input even though they exist at latter positions in a sentence, which is also supported by the relatively higher score of R-1 and R-L.
Although this {\it challenge} caused low readable and ungrammatical summaries, it is an interesting research direction to sophisticate such \ealm's behavior.

\vsecu
\section{Conclusion}
\label{sec:conclusion}
\vsec
We brought the Q-learning framework into unsupervised text summarization and proposed a new method \ealm~that is an edit-based unsupervised summarizer leveraging a Q-learning agent and a language model.
The experments showed that \ealm~performed competitively with the previous encoder-decoder-based methods.
However, in qualitative analysis, we found that the quality of the generated summaries of any unsupervised model was not sufficient, and there are individual limitations for each model.
These issue must be overcome as the step forward to generating practically available summaries without paired data.
In particular for \ealm, there is room for improvement by importing the latest techniques in RL.
Our work paves the way for further research on bridging Q-learning and unsupervised text summarization.


\bibliography{emnlp2020}
\bibliographystyle{acl_natbib}

~
\newpage
~
\newpage
\appendix
\section{Appendices}
\subsection{Autoregressive Prediction with MLM}
\label{sec:autoregressive-prediction-with-MLM}
Algorithm \ref{alg:auto-regressive-predictino-with-mlm} describes the autoregressive prediction with MLM, which we used when an input contains multiple masks.

\begin{algorithm}[htb]
  \caption{{\small Autoregressive prediction with MLM}}
  \label{alg:auto-regressive-predictino-with-mlm}
\begin{algorithmic}
  \STATE {\bfseries Input:} a sentence $\orig$ that includes $\mask$s
  \STATE {\bfseries Outpit:} a sentence $\orig$ after replacing all $\mask$s with predicted words
  \STATE $I \leftarrow (\forall i)\{i \mid x_i = \mask\}$
  \WHILE{$I \neq \phi$}
  \FOR{$j\in I$}
      \STATE $w_j \leftarrow L(\bm{x}_{\backslash j})$ 
  \ENDFOR
  \STATE $j^* \text{ \ } \leftarrow \argmax_{j \in I} P(w_j|x_{\backslash j})$
  \STATE $x_{j^*} \leftarrow w_{j^*}$
  \STATE $I \text{ \ \ } \leftarrow I_{\backslash j^*}$
  \ENDWHILE
\end{algorithmic}
\end{algorithm}

\subsection{Exploration of Prediction Order}
\label{sec:exploration-of-prediction-order}
As explained in section \ref{sec:algorithm} in the main paper, the editorial agent explores the order to predict.
While the agent basically chooses a state with a maximum Q-value as the next state, we sometimes pick a most uncertain state instead.
We define the uncertainty of a state by the entropy of action probabilities as $H(s)= - \sum_{a \in \mathcal{A}} Q(s, a) \log Q(s, a)$, and then $s^*$ and $a^*$ are selected as
\[
    s^* = \argmax_{s \in s^t} H(s) \text{ , } a^* = \argmax_{a \in \mathcal{A}} Q(s^*, a) \text{ .}
\]

\subsection{Semantic Similarity and Log-likelihood Computation in Summary Assessment}
\label{sec:sim-llh}
\paragraph{Semantic Similarity.}
We use a pre-trained model to predict the semantic similarity of paired-sentences with their BERT encodings.\footnote{\url{https://github.com/AndriyMulyar/semantic-text-similarity}}
The model is trained in a supervised manner with a pair of sentences and their similarity score.
The original library outputs a real-valued score in the range of $[0, 5]$, whereas we normalize it to $[0, 1]$.

\paragraph{Log-likelihood.}
We compute the log-likelihood of a compressed sentence by using BERT as follows \cite{wang-cho-2019-bert}:
\[
    \frac{1}{\lcomp} \sum_{i \in \lcomp} \log(P(y_i \mid \comp_{\backslash i})) \text{ .}
\]
However, our $llh$ function performs thresholding --- namely, it returns 1 if the score is beyond a threshold, otherwise 0 --- because the raw log-likelihood score is not scaled with the other rewards.
We empirically set the threshold to 0.005.

\subsection{Relaxations in $rr_{(t)}$ Calculation}
\label{sec:relaxations-reconstruction-rate}
The calculation of the reconstruction rate introduced in section \ref{sec:stepwise-reward-computation} is based on an exact match of each word of $\orig$ and $\reco$.
Given the ambiguity of natural language, this is very strict, so the agent rarely acquires rewards.
We relax this situation by 1) excluding stop words in the calculation and 2) comparing with top-k candidates.
Therefore, the equation of $rr$ can be formally re-written as
\[
    rr_{(t)} = \frac{|\{i \mid x_i \in L^k(\bm{z_{\backslash i}}) \cap x_i \notin \stopwords\}|}{|\{i \mid x_i \notin \stopwords\}|} \nonumber \text{ ,}
\]
where $L^k(\bm{z_{\backslash i}})$ returns top-k probable words for the $i$-th position and $\stopwords$ is a pre-defined set of words.
We set $k = 10$. We used common stopwords (e.g., {\it him}, {\it the}) and infrequent words in $\dgiga$ for $W$.

\subsection{Experimental Details}
\paragraph{Computing Infrastructure.}
We run the models on a machine with the below specifications:

\begin{itemize}
    \item Ubuntu 18.04
    \item Intel(R) Xeon(R) @ 2.60GHz
    \item RAM 120GB
    \item NVIDIA Tesla P100
\end{itemize}

\paragraph{Model Size.}
In \ealm, the number of trainable parameters was 348208 in our experimental setting, which is all for the editorial agent.
There are no trainable parameters for the language model.

\paragraph{Hypperparameter Search.}
We did not conduct a hyperparameter search.
We empirically determined the values described in the main paper (the ``Proposed method" paragraph in $\S$\ref{sec:experiment}).

\paragraph{Runtime Speed.}
\ealm~processes a sentence in three seconds pm average on the above GPU.

\subsection{Generated Summaries}
\label{sec:samples}
Samples of the summaries generated by each model are listed in the tables on the following next pages.
These examples are taken from the first sentences for $\dgiga$ and randomly picked for $\dduca$ and $\dducb$.
We also include human-generated summaries (i.e., gold reference).

\begin{table*}[]
{\small 
\begin{tabular}{p{2.8cm}|p{2.8cm}|p{2.8cm}|p{2.8cm}|p{2.8cm}}
\hline
\multicolumn{1}{c}{INPUT}                                                                                                                                                                                        & \multicolumn{1}{|c}{Human}                                          & \multicolumn{1}{|c}{\seq}                                                                                             & \multicolumn{1}{|c}{\cmatch}                                           & \multicolumn{1}{|c}{\ealm}                                                                                                                               \\ \hline
japan 's nec corp. and UNK computer corp. of the united states said wednesday they had agreed to join forces in supercomputer sales .                                                                            & nec UNK in computer sales tie-up                                   & japan 's nec corp. and her computer corp. of the united states said                                                  & nec agrees to join forces in supercomputer sales                            & nec computer united states said agreed join forces in sales                                                                                            \\ \hline
the sri lankan government on wednesday announced the closure of government schools with immediate effect as a military campaign against tamil separatists escalated in the north of the country .                & sri lanka closes schools as war escalates                          & the sri lankan government on thursday announced the closure of government schools with immediate military country    & sri lankan government announces military campaign against tamil separatists & sri lankan government announced closure government schools effect as military campaign escalated north country                                         \\ \hline
police arrested five anti-nuclear protesters thursday after they sought to disrupt loading of a french antarctic research and supply vessel , a spokesman for the protesters said .                              & protesters target french research ship                             & police arrested five anti-nuclear protesters tuesday after they sought to disrupt her of antarctic protesters        & police arrest five anti-nuclear protesters                                  & police arrested after sought disrupt loading of french antarctic research supply vessel spokesman for said                                             \\ \hline
factory orders for manufactured goods rose \#.\# percent in september , the commerce department said here thursday .                                                                                             & us september factory orders up \#.\# percent                       & factory orders for manufactured goods rose \#.\# percent in september                                                & factory orders rise \#.\# percent in september                              & factory orders manufactured goods rose september commerce said here                                                                                    \\ \hline
the bank of japan appealed to financial markets to remain calm friday following the us decision to order daiwa bank ltd. to close its us operations .                                                            & bank of UNK UNK for calm in financial markets                      & the bank of japan appealed to financial markets to remain calm thursday following decision                           & the bank of daiwa ltd. to close its us operations                           & bank japan appealed financial markets to remain calm following us decision order bank to close us operations                                           \\ \hline
croatian president franjo tudjman said friday croatian and serb negotiators would meet saturday to thrash out an agreement on the last serb-held area in croatia , under a deal reached at us-brokered talks .   & rebel serb talks to resume saturday : tudjman by peter UNK         & croatian president franjo tudjman said thursday croatian and serb negotiators would meet saturday to agreement talks & croatian president franjo tudjman says serb negotiators will meet           & croatian said croatian serb would meet thrash out an agreement on last area croatia under deal reached at talks                                        \\ \hline
japan 's toyota team europe were banned from the world rally championship for one year here on friday in a crushing ruling by the world council of the international automobile federation -lrb- fia -rrb- .     & toyota are banned for a year                                       & japan 's toyota team europe were banned from the world rally championship for one here fia                           & europe is banned from the world championship for one year                   & japan toyota team europe banned from world rally championship for year here in crushing ruling council international automobile .                      \\ \hline
\end{tabular}
\caption{Summaries by Human (gold reference), \seq, \cmatch~and \ealm~from $\dgiga$,}
}
\end{table*}

\begin{table*}[]
{\small 
\begin{tabular}{p{2.8cm}|p{2.8cm}|p{2.8cm}|p{2.8cm}|p{2.8cm}}
\hline
\multicolumn{1}{c}{INPUT}                                                                                                                                                                                        & \multicolumn{1}{|c}{Human}                                          & \multicolumn{1}{|c}{\seq}                                                                                             & \multicolumn{1}{|c}{\cmatch}                                           & \multicolumn{1}{|c}{\ealm}                                                                                                                               \\ \hline
mechanical problems that threaten to shut down the astronomical observations of the hubble space telescope may prompt a repair mission six months earlier than planned to the \$ 1.7 billion spacecraft , nasa officials told congress on wednesday .                                  & Problems may stop Hubble astronomical observations; NASA may accelerate repair mission & mechanical problems that threaten to shut down the her observations of the hubble space telescope threaten .                        & nasa observes                                       & that threaten to shut down astronomical observations of space telescope may prompt repair mission six earlier than planned billion spacecraft nasa officials told congress on           \\ \hline
perhaps no city offers a more public example of the problems of homelessness than san francisco , the biggest complaint visitors lodge about the city concerns the aggressive panhandling and other manifestations of homelessness that they experience , say city tourist officials . & Lack of affordable housing basic to San Francisco's homeless crisis.                   & perhaps no city offers a more public example of the problems of her than san offers more public tourist                             & san francisco city lidge                            & perhaps no city offers more public example of problems of than san francisco , complaint lodge about city concerns aggressive and other of that experience , say city tourist officials \\ \hline
atlanta -- maybe , just maybe , customers who pay to use bank atm machines are beginning to fight back , or maybe they 're just getting smarter .                                                                                                                                      & Bank customers beginning to resist double charges on ATM use.                          & atlanta -- maybe , just maybe , customers who pay to use bank machines getting                                                      & atlanta gets smarter                                & atlanta maybe , just maybe customers who pay to use bank atm machines beginning fight , or maybe getting smarter                                                                        \\ \hline
the head of turkey 's pro-islamic party said thursday he would not insist on his rightful chance to lead turkey 's next government , heading off a confrontation with the military that would only deepen the nation 's political crisis .                                             & Broad-based secularist coalition likely in Turkey.                                     & the head of turkey 's her party said tuesday he would not insist on rightful turkey 's crisis .                                     & the head of turkey 's pro-islamic party             & head of turkey party said he would insist rightful chance to lead turkey next government heading off confrontation with military that would only nation political crisis                \\ \hline
suicide bombers targeted a crowded open-air market friday , setting off blasts that killed the two assailants , injured 21 shoppers and passersby and prompted the israeli cabinet to put off action on the new peace accord .                                                         & Possible early detonation of car bomb still injures 21, bombers killed                 & suicide bombers targeted a crowded open-air market tuesday , setting off blasts that killed assailants accord .                     & israeli cabinet puts off accord on                  & suicide bombers targeted crowded market setting off blasts that killed two , injured 21 and and prompted the israeli cabinet to put off action on new peace accord                      \\ \hline
president nelson mandela acknowledged saturday the african national congress violated human rights during apartheid , setting him at odds with his deputy president over a report that has divided much of south africa .                                                              & President Nelson Mandela acknowledges ANC rights violations. Other leaders disagree.   & president clinton mandela acknowledged saturday the african national congress violated human rights during apartheid setting africa & nelson mandela acknowledges human rights            & mandela acknowledged national congress violated human during setting at odds with deputy president over report that divided much of south                                               \\ \hline
\end{tabular}
\caption{Summaries by Human (gold reference), \seq, \cmatch~and \ealm~from $\dduca$.}
}
\end{table*}

\begin{table*}[]
{\small 
\begin{tabular}{p{2.8cm}|p{2.8cm}|p{2.8cm}|p{2.8cm}|p{2.8cm}}
\hline
\multicolumn{1}{c}{INPUT}                                                                                                                                                                                        & \multicolumn{1}{|c}{Human}                                          & \multicolumn{1}{|c}{\seq}                                                                                             & \multicolumn{1}{|c}{\cmatch}                                           & \multicolumn{1}{|c}{\ealm}                                                                                                                               \\ \hline
endeavour 's astronauts connected the first two building blocks of the international space station on sunday , creating a seven-story tower in the shuttle cargo bay .                                                                                                                                                                                       & First 2 building blocks of international space station successfully joined.  & endeavour 's astronauts connected the first two building blocks of the international space station                                 & endeavour 's astronauts create a shuttle                      & connected first building blocks of international space on creating tower in shuttle bay                                                                                                                                                               \\ \hline
in a cocoon of loyal and wealthy supporters , president clinton said friday that he must `` live with the consequences '' of his mistakes , although he contended that democrats should take pride in the achievements of his presidency and take heart from its possibilities .                                                                             & Clinton supports candidates, speaks at fundraisers, acknowledges mistakes.   & in a her of loyal and wealthy supporters , president clinton said tuesday that must of loyal and wealthy supporters , clinton `` . & democrats take pride in presidency                            & in a of loyal and wealthy supporters president clinton said that must live with consequences of mistakes although he that should pride in achievements of presidency and heart from possibilities                                                     \\ \hline
on the eve of a holiday that has been linked to antiabortion violence , the authorities on tuesday were investigating whether a picture of an aborted fetus sent to a canadian newspaper was connected to last month 's fatal shooting of a buffalo , n.y. doctor who provided abortions or four similar attacks in western new york and canada since 1994 . & Anti-abortion flyer in Canada may be related to Buffalo clinic slaying       & on the eve of a holiday that has been linked to her violence , authorities of holiday that has been linked to her violence ,       & on the eve of a holiday                                       & on eve of holiday that has been linked to violence on investigating picture of an sent to canadian newspaper was connected to last month fatal shooting of buffalo , doctor who provided or similar attacks in western new york and canada since 1994 \\ \hline
famine-threatened north korea 's harvest will be no better this year than last and could be worse , a senior u.n. aid official said saturday .                                                                                                                                                                                                               & World Food Program reports famine may have killed 2 million North Koreans    & her north korea 's harvest will be no better this year than last worse                                                             & south korea 's zhan                                           & north korea harvest better last could worse senior aid official said                                                                                                                                                                                  \\ \hline
matthew wayne shepard , the gay student who was beaten in the dead of night , tied to a fence and left to die alone , was mourned at his funeral friday by 1,000 people , including many who had never met him .                                                                                                                                             & Matthew Shepard eulogized as one who wanted to make people's lives better    & matthew wayne her , the gay student who was beaten in the dead of night , gay student who was beaten him                           & us ceos                                                       & , gay who beaten in dead of night tied to fence and left to die alone was at funeral by including who had met                                                                                                                                         \\ \hline
a delegation of chilean legislators lobbying against the possible extradition of augusto pinochet to spain to face trial , warned thursday that chile was on the brink of political turmoil .                                                                                                                                                                & Chilean legislators protest in Madrid against extradition of Pinochet        & a delegation of chilean legislators lobbying against the possible extradition of augusto pinochet to face turmoil                  & delegation of chilean legislators faces trial                 & delegation of chilean legislators lobbying against possible of augusto spain to trial , warned chile on brink of political turmoil                                                                                                                    \\ \hline

\end{tabular}
\caption{Summaries by Human (gold reference), \seq, \cmatch~and \ealm~from $\dducb$.}
}
\end{table*}

\end{document}